\documentclass[letterpaper, 10 pt, conference]{ieeeconf}
\usepackage[utf8]{inputenc}
\IEEEoverridecommandlockouts
\title{\LARGE \bf Data-Driven Encoding: A New Numerical Method for Computation of the Koopman Operator}
\author{Jerry Ng$^{1}$, H. Harry Asada$^{2}$,$~\IEEEmembership{Fellow,~IEEE}$%
\thanks{This material is based upon work supported by National Science Foundation Grant NSF-CMMI 2021625.} %Use only for final RAL version
\thanks{$^{1}$Jerry Ng is with Department of Mechanical Engineering at the Massachusetts Institute of Technology
        {\tt\small jerryng@mit.edu}}%
\thanks{$^{2}$H. Harry Asada is with Department of Mechanical Engineering at the Massachusetts Institute of Technology
        {\tt\small asada@mit.edu}}%
}
\usepackage{multirow}
\usepackage[ruled, linesnumbered]{algorithm2e}
\usepackage[font=small]{caption}
\usepackage{cite}
\usepackage{amsmath,amssymb,amsfonts}
\usepackage{algorithmic}
\usepackage{graphicx}
\usepackage{textcomp}
\usepackage{xcolor}
\usepackage{mathtools} 
\usepackage{subcaption}
\usepackage{url}
\usepackage{bondgraphs}
\usepackage{standalone}
\begin{document}

\maketitle
\begin{abstract}
This paper presents a data-driven method for constructing a Koopman linear model based on the Direct Encoding (DE) formula. The prevailing methods, Dynamic Mode Decomposition (DMD) and its extensions are based on least squares estimates that can be shown to be biased towards data that are densely populated. The DE formula consisting of inner products of a nonlinear state transition function with observable functions does not incur this biased estimation problem and thus serves as a desirable alternative to DMD. However, the original DE formula requires knowledge of the nonlinear state equation, which is not available in many practical applications. In this paper, the DE formula is extended to a data-driven method, Data-Driven Encoding (DDE) of Koopman operator, in which the inner products are calculated from data taken from a nonlinear dynamic system. An effective algorithm is presented for the computation of the inner products, and their convergence to true values is proven. Numerical experiments verify the effectiveness of DDE compared to Extended DMD. The experiments demonstrate robustness to data distribution and the convergent properties of DDE, guaranteeing accuracy improvements with additional sample points. Furthermore, DDE is applied to deep learning of the Koopman operator to further improve prediction accuracy.

\end{abstract}
% \keywords
% Koopman operator, Dynamic Mode Decomposition, Koopman Direct Encoding \endkeywords
\section{Introduction}
\label{intro}

%The original paper regarding the Koopman Operator was written in 1931, detailing 

Dynamic Mode Decomposition (DMD) was presented as a method to produce linear models from data generated through nonlinear dynamical processes by using Singular Value Decomposition (SVD) \cite{schmid2010dynamic}. Later, this method was developed further to create Extended Dynamic Mode Decomposition (EDMD), which introduced the concept of using observable functions, nonlinear functions of state variables, as a method of augmenting the state space \cite{williams2015data}. EDMD referenced the Koopman Operator as justification and a theoretical underpinning for lifting the state space. In his seminal work, Bernard Koopman showed the existence of this operator that transforms nonlinear systems into linear systems \cite{koopman1931hamiltonian}. Another extension of DMD has shown the viability of using DMD for control on non-autonomous systems \cite{proctor2016dynamic}. This enabled complex nonlinear Model Predictive Control (MPC) to be converted to linear MPC \cite{Korda_2018}, leading to numerous studies  utilizing the methodology to real systems\cite{ng2021model, cibulkaVehicle2020, calderon2021, bruder2021}. %This significant body of research, an accumulation of fifteen years of work, has been reviewed and summarized \cite{schmid2022review}. 
 
To improve the accuracy of the models based on the Koopman Operator, two avenues of research have formed. The first avenue regards the selection of the observable functions used for constructing a lifted state space, as these functions are a key ingredient in creating an accurate linear model. Various methods have been developed, including deep neural networks for learning effective observable functions \cite{Lusch_2018, yeung2017, Selby_2021} and optimization \cite{korda2020optimal}. However, an efficient selection of observables does not solve all the issues that arise when attempting to construct an accurate linear model. For example, it is known that unstable modes are involved in Koopman-based DMD models and their extensions although the underlying nonlinear systems are known to be stable. The second avenue of research involves the formulation of the linear transition matrix. Extensive studies have been done to create stable linear models to remedy the situations where an outright use of DMD would lead to the creation of an unstable linear model \cite{fanstabkoop2021, mamakoukas2020learning, bevanda2022diffeomorphically, han2021desko}. Recently, an extension of DMD, called Robust Dynamic Mode Decomposition (RDMD), utilizes statistical measures to suppress the effect of outliers on modeling the linear Koopman matrix\cite{Hossein_Abolmasoumi_2022}. %This second avenue of research, concerning the method of constructing the linear matrix, is the subject of the current work.

A fundamental difficulty in constructing a proper linear model is data dependency. %; inconsistent estimates of model parameters may result, depending on a data sample. 
Least Squares Estimation (LSE), involved in all DMD based methods, often produces a significant bias due to the distribution of the dataset. To eliminate this dependency on data distributions, the current work takes an alternative approach to LSE.

Recently, a new formulation of the Koopman Operator, termed Koopman Direct Encoding (DE), was produced \cite{asada2022}. This method directly encodes the nonlinear function of state transition using observables as basis functions to obtain a Koopman linear model. Inner products of observable functions in composition with the nonlinear state transition function are used to construct the state transition matrix without use of LSE. While DE theoretically guarantees the exact linear model, it requires access to the nonlinear state equations, which are often not available in practical applications. The current work aims to fill the gap between DE and data-driven approaches.

%There is not a simple conversion between the DMD methods and DE, even while utilizing the same observable functions, for many reasons. One of which is because DE utilizes the underlying nonlinear model in combination with the observable functions to generate the linear matrix. In addition, the methods are calculated in fundamentally different manners; DE utilizes integration over the state space, whereas DMD methods utilize least squares estimation based on data. However, the need for the underlying nonlinear model is a significant limitation for DE. Thus, there is a gap between DE and a method that can be utilized for real applications that are currently able to be handled with DMD as it is a data-driven method. 

There are four significant contributions presented in this work. The first is the conversion of the DE formula of the Koopman Operator to a data-driven formula. The second is a computational algorithm and proof of its convergence to the true inner products that constitute the DE formula. The third is numerical experiments that provide evidence that the proposed method, unlike EDMD, does not exhibit biases to data distribution, but can produce consistently higher accuracy compared to EDMD. %more accurate models thanks to the new formulation. Models created through this novel method to EDMD models utilizing identical observable functions, providing numerical evidence of these claims. 
Finally, the DDE algorithm is utilized in modeling a high order nonlinear system in combination with deep learning.

%The paper is outlined as follows. In Section \ref{preliminaries}, we provide an overview of the Koopman Operator, formulated with both approaches. Then, in Section \ref{methods}, the discussion comparing the two methodologies is presented; afterwards, the novel numerical integration approach is articulated. We then present a series of results comparing the novel method to EDMD, along with discussion of observations made by the authors in Section \ref{experiments}.  Concluding remarks are offered in Section \ref{conclusion}.
\section{Koopman Operator and the Direct Encoding Method}
\label{preliminaries}

In this section, we give a brief overview of the Koopman Operator and dynamic mode decomposition, and introduce the direct encoding method for obtaining a Koopman operator directly from nonlinear dynamics.

\subsection{Least Squares Estimation of the Koopman operator}

Consider a discrete-time dynamical system, given by

%The basis of this linearization is to create a linear system composed of a larger state space than required to represent the original nonlinear system. In discrete time, it begins with a nonlinear dynamic equation
\begin{equation}
    x_{t+1} = f(x_t) 
    \label{eq:gen_nonlinear}
\end{equation}
where $x \in X \subset \mathbb{R}^n$ is the independent state variable vector representing the dynamic state of the system, 
$f$ is a self-map, nonlinear function $f : X \rightarrow X$, and $t$ is the current time step. Also consider a real-valued observable function of the state variables $g : X \rightarrow \mathbb{R}$. The Koopman Operator $\mathcal{K}$ is an infinite-dimensional linear operator acting on the observable function $g$ :
\begin{equation}
    \mathcal{K} g  = g \circ f
\end{equation}
where $g \circ f$ is the composition of function $g$ with function $f$: $(g \circ f)(x) = g(f(x))$. %The Koopman operator is linear, even though the dynamic system is nonlinear. 

A common data-driven method for constructing the operator is Extended Dynamic Mode Decomposition (EDMD) \cite{williams2015data}, where observables that are experimentally obtained or simulated from the governing equation of the system are augmented by including real-valued observable functions of the independent state vector $x_t$. Collectively, a high-dimensional state vector is formed. 
\begin{equation}
    z_t = \begin{bmatrix} g_1(x_t) \\ g_2(x_t) \\ \vdots \\ g_m(x_t) \end{bmatrix}
\end{equation}
where $m$ is the order of the lifted state corresponding to the number of observable functions. Underpinned by the Koopman Operator theory, EDMD assumes the existence of a linear state transition matrix $A$ relating $z_{t+1}$ to $z_t$, and determine $A$ by solving a least squares regression that minimizes the Sum of Squared Error (SSE).

%allowed to be any  function of the state variable. It dictates that an approximation of the Koopman Operator, $A$, can be found through a least squares regression that minimizes
\begin{equation}
   A  = \arg \min_A \sum_{t}|| z_{t+1} - A z_t || ^2
\end{equation}
Singular Value Decomposition (SVD) is used for the least squares optimization.
% \begin{equation}
% \phi_i \in \mathcal{H}
% \end{equation}
\subsection{Direct Encoding of the Koopman Operator}
\label{koop:de}

An alternative to the least squares estimate and EDMD is to obtain the exact $A$ matrix by directly encoding the self-map, nonlinear state transition function $f(x)$ with an independent and complete set of observable functions through inner product computations. This Direct Encoding method is introduced next, while the full proof can be found in \cite{asada2022}. 
%Our prior work has demonstrated the ability to directly encode the Koopman Operator from the system of equations that govern the dynamic system. An abbreviated formulation is written here, 

%Let $[\phi_1, \phi_2, \phi_3, ...]$ be orthonormal basis functions spanning a Hilbert space $\mathcal{H}$. Any observable $g_i$ in the Hilbert space, $g_i \in \mathcal{H}$, can be expanded in the orthonormal basis as:

%\begin{equation}
%    g_i = \sum_{j=1}^\infty \langle g_i, \phi_j \rangle \phi_j 
%    \label{eq:g_i expansion}
%\end{equation}

%Assuming that the self-map nonlinear function $f(x)$ is continuous, we take the composition of $g_i$ and $\phi_j$ with $f(x)$ on both sides of eq. (\ref{eq:g_i expansion}).

%\begin{equation}
%    g_i \circ f = \sum_{j=1}^\infty \langle g_i, \phi_j \rangle (\phi_j \circ f)
%    \label{eq:obsf_exp}
%\end{equation}

Let us first consider the case where $g_1, g_2, g_3, ...$ are orthonormal basis functions spanning a Hilbert space $\mathcal{H}$.
We assume that the self-map nonlinear function $f(x)$ is continuous and that the composition of $g_j$ with $f$ is also involved in the Hilbert space.
\begin{equation}
    g_j \circ f \in \mathcal{H} \quad\forall j
\end{equation}
%and $[g_1, g_2, g_3, ...]$ also form a set of orthonormal basis functions spanning the Hilbert space. Then,
This implies that the function $g_j \circ f$ can be expanded in $[g_1, g_2, g_3, ...]$.
\begin{equation}
    g_j \circ f = \sum_{k=1}^\infty \langle g_j \circ f, g_k \rangle g_k
    \label{eq:orthof_exp}
\end{equation}

%Substituting eq. (\ref{eq:orthof_exp}) to eq. (\ref{eq:obsf_exp}) yields
%\begin{equation}
%    g_i \circ f = \sum_{j=1}^\infty \langle g_i, \phi_j \rangle \sum_{k=1}^\infty \langle \phi_j \circ f, g_k \rangle g_k
%    \label{eq:g-f double sum}
%\end{equation}

Concatenating $g_1, g_2, g_3, ...$ and $g_1 \circ f, g_2 \circ f, g_3 \circ f ...$ in infinite dimensional column vectors, respectively,
\begin{equation}
   \bar{z}_t = \begin{bmatrix} g_1(x_t) \\ g_2(x_t) \\ \vdots  \end{bmatrix},
   \bar{z}_{t+1} = \begin{bmatrix} g_1[f(x_t)] \\ g_2[f(x_t)] \\ \vdots  \end{bmatrix}
\end{equation}
eq. (\ref{eq:orthof_exp}) can be written in matrix and vector form.
\begin{equation}
    \bar{z}_{t+1} = \bar{A} \bar{z}_t
    \label{eq:X_t+1 = A X_t}
\end{equation}
where $\bar{A}$ is an infinite dimensional matrix consisting of the inner products involved in eq. (\ref{eq:orthof_exp}),

\begin{equation}
    \bar{A} = \begin{bmatrix} \langle g_1\circ f, g_1 \rangle & \langle g_1 \circ f, g_2 \rangle & \dots \\ \langle g_2 \circ f, g_1 \rangle & \langle g_2 \circ f, g_2 \rangle & \dots \\ \vdots & \vdots & \ddots \end{bmatrix}
\end{equation}
%\begin{equation}
%    \bar{A}= \sum_{j=1}^\infty \begin{bmatrix} \langle g_1, \phi_j \rangle \\ \langle g_2, \phi_j \rangle \\ \vdots\end{bmatrix} \begin{bmatrix} \langle \phi_j \circ f, g_1 \rangle & \langle \phi_j \circ f, g_2 \rangle & \dots \end{bmatrix}
%\end{equation}
Eq. (\ref{eq:X_t+1 = A X_t}) manifests that the state lifted to the infinite dimensional space $\bar{z_t}$ makes linear state transition with matrix $\bar{A}$.

The observables $g_1, g_2, g_3, ...$ were assumed to be orthonormal basis functions in the above derivation. This assumption can be relaxed to an independent and complete set of basis functions spanning the Hilbert space. Hereafter, let $[g_1, g_2, g_3, ...]$ be an independent and complete set of basis functions spanning the Hilbert space. 

%Using the orthonormal basis functions $[\phi_1, \phi_2, \phi_3, ...]$, each observable can be expanded to $g_i = \sum_{j=1}^\infty \langle g_i, \phi_j \rangle \phi_j$. This implies that there is a linear relationship between $g_i's$ and $\phi_j's$.
%\begin{equation}
%    \begin{bmatrix} g_1 \\ g_2 \\ \vdots \end{bmatrix} = C \begin{bmatrix} \phi_1 \\ \phi_2 \\ \vdots \end{bmatrix} 
%    \label{eq:map}
%\end{equation}
%where
%\begin{equation}
%    C = \begin{bmatrix} \langle g_1, \phi_1 \rangle & \langle g_1, \phi_2 \rangle & \hdots \\ \langle g_2,\phi_1 \rangle & \langle g_2, \phi_2 \rangle & \hdots \\ \vdots & \vdots & \ddots \end{bmatrix}
%\end{equation}
%Note that, as long as $[g_1, g_2, g_3, ...]$ span the Hilbert space, the above matrix $C$ is non-singular and therefore invertible. This allows us to extend eq. (\ref{eq:X_t+1 = A X_t}) to the one in terms of the independent and complete set of observables.

It can be shown that the time evolution of lifted state $z_t$ is given by a constant matrix $A_f$ for the independent and complete set of basis functions $[g_1, g_2, g_3, ...]$.
\begin{equation}
    z_{t+1} = A_f z_t
    \label{eq:A_f 1}
\end{equation}
%where the two matrices are related as

%\begin{equation}
%    A_f = C \bar{A}C^{-1}
    \label{A_f with C}
%\end{equation}

%Notably, once the existence of the linear state transition is guaranteed in eq. (\ref{eq:A_f 1}), the matrix $A_f$ can be obtained without use of the orthonormal basis functions $[\phi_1, \phi_2, \phi_3, ...]$ and the matrix $C$ and its inverse. 
The matrix $A_f$ can be computed directly from the self-map, state transition function $f(x)$ and an independent and complete set of observables $[g_1, g_2, g_3, ...]$ through inner product computations. Post-multiplying the transpose of $z_t$ to both sides of eq. (\ref{eq:A_f 1}) and integrating them over $X$ yield:
\begin{equation}
    \int_X z(f(x)) z^T(x)dx = A_f \int_X z(x) z^T(x) dx
\end{equation}
which can be written as
\begin{equation}
    Q = A_f R
\end{equation}
where
\begin{align}
    Q &= \begin{bmatrix} \langle g_1\circ f, g_1 \rangle & \langle g_1 \circ f, g_2 \rangle & \dots \\ \langle g_2 \circ f, g_1 \rangle & \langle g_2 \circ f, g_2 \rangle & \dots \\ \vdots & \vdots & \ddots \end{bmatrix}\\
    R &= \begin{bmatrix} \langle g_1, g_1 \rangle & \langle g_1, g_2 \rangle & \dots \\ \langle g_2, g_1 \rangle & \langle g_2, g_2 \rangle & \dots \\ \vdots & \vdots & \ddots \end{bmatrix}
\end{align}
Because the observables are independent, the matrix $R$ is non-singular. Therefore, the matrix $A_f$ is given by
\begin{equation}
    A_f = QR^{-1}
    \label{eq:A_f = Q*R_inv}
\end{equation}

This formula for obtaining the matrix $A_f$ directly from the governing nonlinear state equation with the function $f(x)$ and the independent observables through inner products, which are guaranteed to exist in Hilbert space $\mathcal{H}$, is the Direct Encoding method.

\section{Data-Driven Koopman Encoding}
\label{methods}

The prevailing method for construction of the Koopman Operator, EDMD, is based on LSE and SVD. This method, however, cannot provide a consistent estimate; the result is highly dependent on the distribution of data as the Koopman Operator is being approximated \cite{ljung1998system}. This dependency on distribution occurs because a core assumption of LSE is that the model structure is correct; when this assumption is violated, LSE is unable to create an unbiased estimator \cite{freedman2009statistical}. As the Koopman operator is truncated in practical use, this assumption does not hold.
%If data are densely populated in a particular region, LSE provides an estimate biased towards the dense region. 
%This occurs despite the possibility that a region low in data-density region can exhibit a uniquely different spectrum of dynamics from the high density region. 

Non-uniform data distributions inevitably occur in practical applications. For a nonlinear dynamical system with a stable equilibrium, for example, data collected from experiments and/or simulation of the system tend to be dense in the vicinity of the equilibrium, as all trajectories that begin within a region of attraction converge to the equilibrium. %that is illustrated in Fig. \ref{fig:illus}, in which a situation where a large number of points is measured in a small dynamic region, and points away from the densely populated region are sparse. 
Because LSE applies equal weighting to all data points, the model is heavily tuned to the behavior of the densely populated region.
%Of course, this dilemma does not arise when the observable functions are completely and correctly chosen as the Koopman Operator can be exactly determined in that case, but this is not the general case in applications of the Koopman Operator. 

% Another limitation of EDMD based on LSE is the limited insight gained from application of the method. The prediction error that arises from utilizing the method comes from two sources. The first source is the set of observables chosen to lift the space, and the second is the data-dependent nature of the inconsistent estimate. LSE does not explain which is the cause of poor estimation performance. 

The Direct Encoding method described previously enables us to obtain the exact linear state transition matrix $A$ through inner product computations. As the formulation is based on integration over the entire state space, there is no bias towards particular parts of the domain. 
%As long as proper observables are chosen and the inner products are computed properly, a legitimate result is obtained. 

However, the original form of the Direct Encoding method utilizes the nonlinear state equation, i.e. the self-map $f(x)$, to compute the inner products. In practical applications, such a nonlinear function is not always available; only data are available. The objective of this section is to establish a computational algorithm to obtain the $A$ matrix by  numerically computing the inner products, $\langle g_i, g_j \rangle, \langle g_i \circ f, g_j \rangle$, from a given set of data. 

The method presented consists of three operations. 
\begin{itemize}
    \item The integral of the inner products is reduced in range from the entire state space to the dynamic range encapsulated by the data.
    \item The dynamic range is discretized with data points.
    \item The inner product integral is reduced to a weighted summation of the integrand evaluated at each data point multiplied by the volume $\Delta v$ associated to each point.
\end{itemize}
 Naturally, if data are densely populated in a small region, the discretized integral interval is small and thereby the volume also becomes small. Similarly, the volume tends to be larger where the data are sparse. In the summation, the integrand evaluated at individual data points are "weighted" by the size of the volume. This numerical inner product calculation prevents overemphasis of clustered data.

\subsection{Inner Product Computation}

We present the data-driven encoding method (DDE) as an alternative data-driven method to DMD for calculating a finite order approximation of the Koopman Operator. %In the algorithm presented, we utilize a mesh generation algorithm, such as Delaunay Triangles \cite{cheng2013delaunay} for an $n$-dimensional nonlinear system, where $n \geq 2$. %Depending on the dimension of the system, different methods of mesh construction can be used interchangeably with Delaunay Triangles.
% The $A_\infty$ matrix produced in real applications of DE at finite orders can be seen as finite order projections of the correct infinite order linear system. Mathematically, we have an $A_\infty$ where
% \begin{align}
%     R_\infty = \begin{bmatrix} R & \cdots \\ \vdots & \ddots \end{bmatrix} \\
%     Q_\infty = \begin{bmatrix} Q & \cdots \\ \vdots & \ddots \end{bmatrix} \\
%     A_\infty = Q_\infty R_\infty^{-1}
% \end{align}
% where $\cdots$ is a placeholder for values that are not included in our finite order estimation. These $~$ values are effectively set to zero in our estimation, and thus seems somewhat intuitive to argue that this is a low order projection of the true $A_\infty$ matrix. The "projection" matrix in this case would be
% \begin{equation}
%     P = \begin{bmatrix}I & 0 \\ 0 & 0 \end{bmatrix}
% \end{equation}
% where $I$ is the identity matrix of the corresponding size and
% \begin{align}
%     R = P R_\infty \\
%     Q = P Q_\infty
% \end{align}
% and the zeros are truncated from the lower order matrices $R$ and $Q$. 
The objective of this method is to compute the matrices $R$ and $Q$ in (13) and (14) from data. This entails the computation of inner products:
\begin{align}
    \langle g_i, g_j \rangle = \int_X G_{ij}(\xi)  d\xi \\
    \langle g_i \circ f, g_j \rangle = \int_X F_{ij}(\xi)  d\xi
\end{align}
where
\begin{equation}
    G_{ij} (x) = g_i(x) \bar{g}_j(x),  F_{ij} (x) = g_i[f(x)] \bar{g}_j(x)
\end{equation}
are assumed to be Riemann Integrable; the functions are bounded and continuous \cite{davis2007methods}.

\begin{figure} [!h]
    \centering
    \includegraphics[width = 0.8\linewidth]{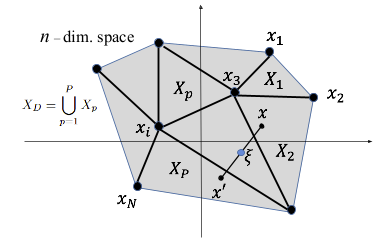}
    \caption{Illustration of the dynamic range of a dataset defined by the convex hull containing all points in the set, partitioned using a triangulation method. The data points are in black and the convex hull that encapsulates all data points is in grey.}
    \label{fig:convexhull}
\end{figure}

There are two data sets used for the inner product computation. The first data set is
\begin{equation}
    D_N = \{ x_i \mid i = 1, \cdots, N; x_i \in X \} 
\end{equation}
Note that all the data values are finite, $|x_i| < \infty$. As such, the integral interval of the inner products is finite in computing them from the data. To define the integral interval, we consider the dynamic range of the system, $X_D$, determined from the data set $D_N$. See Fig.\ref{fig:convexhull}. The dynamic range $X_D$ is defined to be the minimum domain in the space $X$ that includes all the data points in $D_N$, $X_D \supset D_N$, and that is convex. Namely, for any two states in $X_D$, $x, x' \in X_D$,
\begin{equation}
    \xi = \alpha x + (1-\alpha) x' \in X_D
\end{equation}
where $0 \leq \alpha \leq 1$.  %Note that $X_D$ is the smallest domain that is convex and that includes all the data points, as shown in Fig. \ref{fig:convexhull}. 
Each data point $x_i$ is mapped to $f(x_i)$, following the state transition law in eq.(\ref{eq:gen_nonlinear}). We assume that the transferred state, too, stays within the same dynamic range $X_D$. Collecting all the transferred states yields the second data set.
\begin{align}
    D_N^f = \{ f(x_i) \mid i = 1, \cdots, N; x_i \in D_N \} \\
    D_N^f \subset X_D
\end{align}
This implies that the state space of the nonlinear system under consideration is closed within the dynamic range $X_D$.

\begin{figure}
    \centering
    \includegraphics[width = 0.8\linewidth]{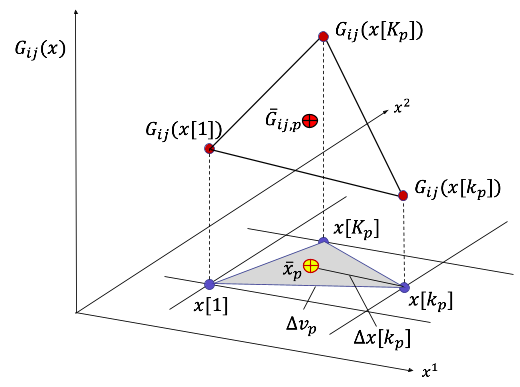}
    \caption{Visualization of the integrand calculation process. The volume of the partition is encapsulated by the data points is denoted as $\Delta v_p$. With this grouping, the value of $G_{ij}$ is calculated for each point and the average among this group is computed, $\bar{G}_{ij,p}$.  In turn, this value, weighted by the volume of this partition, is summed across other partitions (not shown) to approximate the value of the element $R_{ij}$.}
    \label{fig:partition}
\end{figure}

With this dynamic range, we redefine our objective to compute the inner products over $X_D$.
\begin{align}
    R_{ij} = \int_{X_D} G_{ij}(x) dx \label{eq:G_ij integral} \\
    Q_{ij} = \int_{X_D} F_{ij}(x) dx \label{eq:F_ij integral}
\end{align}
The integrals can be computed by partitioning the domain $X_D$ into many segments $X_1, \cdots X_P$, as shown in Fig. \ref{fig:convexhull}.
\begin{equation}
    X_D = \bigcup_{p=1}^P X_p
\end{equation}
We generate these segments by applying a meshing technique to the data set $D_N$, where the $n$-dimensional coordinates of individual data points are treated as nodes of a mesh. Delaunay Triangulation, for example, generates a triangular mesh structure with desirable properties \cite{cheng2013delaunay}. As illustrated in Fig. \ref{fig:convexhull}, each triangular element is convex and has no internal node. The volume of the dynamic range $V(X_D)$ is the sum of the volumes of all the elements.
\begin{equation}
    V(X_D) = \sum_{p=1}^P \Delta v_p
\end{equation}
Accordingly, the integral $R_{ij}$ in eq.(\ref{eq:G_ij integral}) can be segmented to
\begin{equation}
    R_{ij} = \sum_{p=1}^P \int_{X_p} G_{ij} (x) dx
\end{equation}
Suppose that the $p$-th element has $K_p$ nodes, as shown in Fig.\ref{fig:partition}. Renumbering these nodes 1 through $K_p$,
\begin{equation}
    \{ x[k_p] \Big| x[k_p] \in X_p; k_p = 1, \cdots, K_p \},  p = 1, \cdots, P
\end{equation}
The integrand $G_{ij}$ within the $p$-th element can be approximated to the mean of the $K_p$ nodes involved in the $p$-th element.
\begin{equation}
    G_{ij} (x;p) \approx \bar{G}_{ij,p}= \frac{1}{K_p} \sum_{k_p = 1}^{K_p} G_{ij}(x[k_p]), x[k_p] \in X_p
    \label{integrand_approx}
\end{equation}
If Delaunay Triangulation is used, $K_p = n+1$.
See Fig. \ref{fig:partition}. Substituting this into (\ref{eq:G_ij integral}) yields the approximate value of $R_{ij}$.
\begin{equation}
    \hat{R}_{ij} = \sum_{p=1}^P \frac{1}{K_p} \sum_{k_p = 1}^{K_p} G_{ij}(x[k_p])  \Delta v_p
    \label{eq:r_calc}
\end{equation}
where 
\begin{equation}
     \Delta v_p = \int_{X_p} 1 \cdot dx
\end{equation}

Similarly, each component of the matrix $Q$ can be computed by using the same meshing.
\begin{equation}
    \hat{Q}_{ij} = \sum_{p=1}^P \frac{1}{K_p} \sum_{k_p = 1}^{K_p} F_{ij}(x[k_p]) \Delta v_p
    \label{eq:q_calc}    
\end{equation}
Note that $F_{ij}$ is evaluated by using the data points in both $D_N^f$ and $D_N$,
\begin{equation}
    F_{ij}(x[k_p]) = g_i[f(x[k_p])] \bar{g}_j(x[k_p])
\end{equation}
where $f(x[k_p]) \in D_N^f$.
%Fig. \ref{fig:partition} visualizes this process of calculating $G_{ij}$ for a set of data points that encapsulate a single partition of the state space. 

% in general.

\subsection {Convergence}

Consider the center of each partition, $\bar{x}_p = \int_{X_p} x dx / \Delta v_p$, and the distance between $\bar{x}_p$ and each point, $x[k_p]$:
\begin{equation}
    \Delta x[k_p] = \bar{x}_p - x[k_p]
\end{equation}
See Fig.\ref{fig:partition}. The maximum distance from the center of the partition to each point that makes up the partition is 
\begin{equation}
    |\Delta x_p| = \max\{|\Delta x[1]|,\, ... \, |\Delta x[k_p-1]|, \, |\Delta x[k_p]| \}
\end{equation}
Consider a sequence of refining the approximate inner product integral $\hat{R}_{ij}$ by increasing data points $N$.  We can show that, as the number of partition $P$ tends infinity and the maximum subintervals $|\Delta x_p|$ approach zero, the approximate inner product integral $\hat{R}_{ij}$ converges to its true integral.  

%Eq. (\ref{integrand_approx}) converges to the true integrand for the partition for
\begin{equation}
    R_{ij} =  \lim_{\substack{ P \to \infty \\ |\Delta x_p| \to 0}} \ \sum_{p=1}^P \frac{1}{K_p} \sum_{k_p = 1}^{K_p} G_{ij}(x[k_p])  \Delta v_p
\end{equation}
    
    %\frac{1}{K_p} \sum_{k_p = 1}^{K_p} G_{ij}(x[k_p]), x \in X_p
This formulation takes the form of weighted sums, specifically Riemann sums. %This summation is formulated as a sequence that can be refined, where the maximum subintervals, in this case $\Delta x_p$, approach 0. 
Given functions that are bounded and continuous over the subdomain of interest, sequences of this form are known to have a common limit and thus converge upon refinement to the Riemann integral value over that subdomain, according to Numerical Integration theory \cite[Section 1.5]{davis2007methods}. %, section 1.5 of  This convergence can be observed in the refinement of $\Delta x_p$ and more broadly in the refinement of $\Delta v_p$ as the values approach 0. 

\subsection{Algorithm}

In the prior section, integrals (\ref{eq:r_calc}) and (\ref{eq:q_calc}) are presented as summations over partitions. This computation can be streamlined by converting the summations over partitions to the one over nodes. Consider node 3 associated to data point $x_3$ in Fig.\ref{fig:convexhull}, for example. This node is an apex of the 5 surrounding triangles. This implies that integrand $G_{ij}(x)$ is calculated or recalled 5 times in computing (\ref{eq:r_calc}) and (\ref{eq:q_calc}). This repetition can be eliminated by computing volume $\Delta v_k$ associated to node $k$, rather than partition $p: \Delta v_p$. Namely, we compute 
\begin{equation}
    \Delta v_k = \sum_{p=1}^{P} \frac{\Delta v_p}{K_p}I(k,p)
\end{equation}
where $I(k,p)$ is a membership function that takes value 1 when node $k$ is an apex of triangle $p$, that is, node $k$ is involved in partition $p$. Using this volume as a new weight we can rewrite (\ref{eq:r_calc}) and (\ref{eq:q_calc}) to be
\begin{align}
    \hat{R}_{ij} = \sum_{k=1}^K G_{ij}(x[k])  \Delta v_k  \label{eq:r_calc2}\\
    \hat{Q}_{ij} = \sum_{k=1}^K F_{ij}(x[k]) \Delta v_k  \label{eq:q_calc2}
\end{align}

%is the number of partitions that the node $k$ is a simplex of. 
%This creates the situation where the processing of data is handled simultaneously with choosing observable functions. In this subsection, we present 
Using this conversion, the computation can be streamlined and cleanly separated into three steps, as shown by pseudo-code in Algorithm \ref{algo:dde}.  %a three step algorithm that enables the processing of data to be separated from the selection of the observable matrices. 
The steps are: (1) \textit{Graph Creation:} data are connected to create partitions of the domain using a mesh generator: lines 3 to 8, (2) \textit{Weighting Calculation:} calculation of the weights for each data point: lines 10 to 17, and (3) \textit{Matrix Calculation:} the calculation of the $R$ and $Q$ matrices to find the  matrix $A$, lines 19 to 21. %Algorithm \ref{algo:dde} shows the pseudo code of this procedure. performs a summation over the $K$ nodes of the dataset rather than partitions, where each node $k$ corresponds to a point $x_t$, $x_{t+1}$ and is associated with a volume $\Delta v_k$. In this alternative, the same initial steps of the DDE algorithm are followed, noted in lines 3-8 of Algorithm, where the dataset associates the state at one time step with the following time step and used to partition the state space. Afterwards, in lines 10-17, volumes associated to nodes are calculated. Based on (\ref{eq:r_calc}) and (\ref{eq:q_calc}), these volumes are found for each node $k$ via

\begin{algorithm} [!b]
    \caption{Algorithm for DDE in pseudocode}
    \label{algo:dde}
    \KwIn{}
    $D_N$, $D_N^f$\\ 
    \KwOut{}
    $P$: $p$, $K$: $k$, $R$, $Q$ $A$\\
    \textit{Graph Creation:} \\
    \For{$x_i$ in $D_N$ and corresponding $f(x_i)$ in $D_N^f$}{
    Create node, $k$\\
    Assign node attributes for current data point $k.x_t = x_t$ and $k.x_{t+1} = x_{t+1}$  \\
    Append node to node list $K$.
    }
    \textbf{then}{ \\
    \textit{Weighting Calculation:} \\
    Create list of triangles, $P$ using Delaunay Triangulation on node list $K$ using $k.{x_t}$}\\
    \For{$p$ in $P$}{
    Calculate volume, $V$, of $p$\\
    \For{$k$ corresponding to $K_p$}{
    Update volume of each node $k.\Delta v_k = k.\Delta v_k + \frac{V}{n+1}$}
    }
    \textbf{then}{\\
    \textit{Matrix Calculation:}\\
    Find $R$ and $Q$ via (\ref{eq:r_calc2}) and (\ref{eq:q_calc2})\\
    Find $A = Q R^{-1}$
    }
\end{algorithm}

\section{Experiments}
\label{experiments}

\begin{figure} 
    \centering
    \includegraphics[width = 0.9\linewidth]{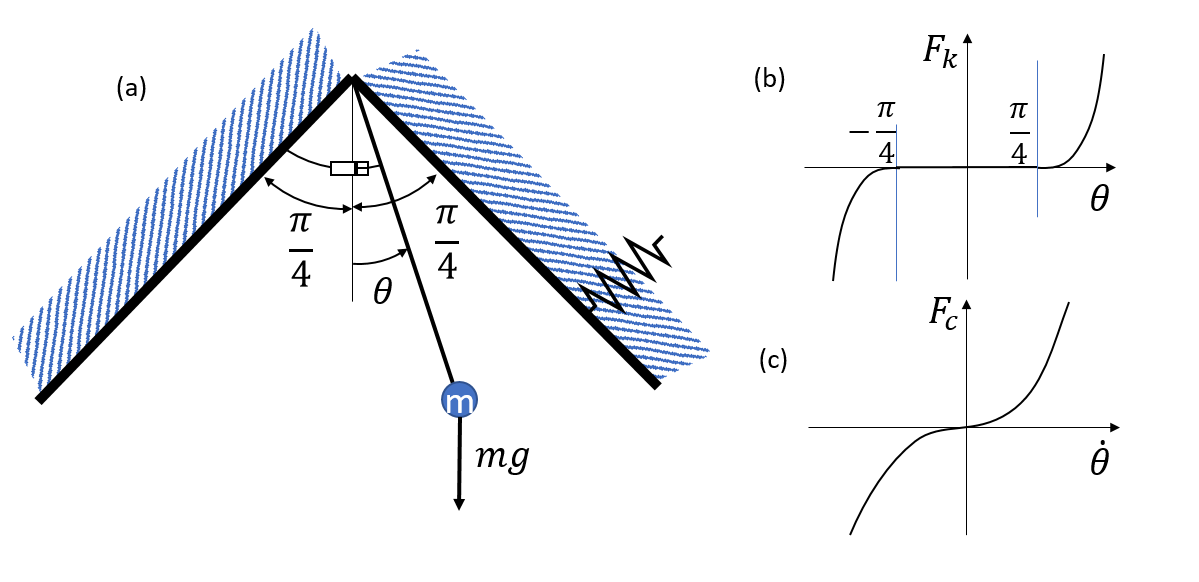}
    \caption{Diagram of pendulum with with walls. (a) depicts the range of the pendulum where the walls are equally angularly displaced from the vertical. (b) depicts the forces exerted on the pendulum due to contact with walls. (c) depicts the damping force exerted onto the pendulum. }
    \label{fig:penddiagram}
\end{figure}

In this section, the DDE algorithm is implemented for the sake of evaluating its validity and comparing its modeling accuracy to EDMD.
%In the following experiments, both EDMD and DDE models are composed with the same observable functions. 
%The goal of the experiment is to demonstrate the differences between DDE and DMD models for two different data sets. 
%The goal of the first experiment, the pendulum experiment, is to demonstrate the differences between DDE and DMD models for datasets with different distributions of points. Notably, the two types of datasets used do have two different dynamic ranges and, as such, the models produced utilizing one dataset are not compared to models produced utilizing the other dataset. 
%The second experiment, the winch experiment, demonstrates the difference between prediction models with a practical high order application that utilizes a neural network to generate observable functions.
%\subsection{Pendulum}
Consider a 2nd order nonlinear system consisting of a pendulum with a nonlinear damper. See Fig. \ref{fig:penddiagram}. The pendulum also bounces against  walls with nonlinear compliance. The state variables for this system are $x = [ \theta, \dot{\theta}]^T$, and the equation of motion can be written as:
%\begin{equation}
%    x = \begin{bmatrix} \theta \\ \dot{\theta} \end{bmatrix}
%\end{equation}
\begin{equation}
    \ddot{\theta} = - sin(\theta) + F_k + F_c
\end{equation}
where $F_k$ and $F_c$ are wall reaction moment and damping moment, respectively,
\begin{align}
    F_k &= \begin{cases}-\text{sign}(\theta) \: k(|\theta| - \frac{\pi}{4})^2 &\text{if }|\theta| \geq \frac{\pi}{4}\\
    0 &\text{if }|\theta| < \frac{\pi}{4} 
    \end{cases}\\
    F_c &= -\text{sign}(\dot{\theta}) \: c \:\dot{\theta}^2
\end{align}
where $k = 200$ and $c = 1$. 
We present a two part numerical experiment for this system. The first experiment regards variations in dataset size and distribution, and the second experiment varies the usage of observable functions.

%A diagram of this physical system is presented in Fig. \ref{fig:penddiagram}.

\subsection{Dataset Variations}

The datasets tested are of three types:
\begin{enumerate}
\item \textbf{Uniform}: These datasets are composed of a rectangular dynamic range which is sampled uniformly, like an evenly divided grid. The range varies from $\theta = [-
0.8,0.8]$ and $\dot{\theta} = [-2,2]$, where the mass can hit the walls and the damping can vary from 0 to a significant value. See Fig. \ref{fig:penddiagram}-(b), (c). %chosen such that all nonlinearities of the system are active in the dataset.
\item \textbf{Gaussian}: Data points are sampled with a finite-support Gaussian distribution. The data are distributed non-uniformly with their highest density at the peak of the Gaussian placed at diverse locations. In addition, each dataset contains 100 data points uniformly distributed along the boarder of the dynamic range to guarantee the same dynamic range as the uniform datasets.  Samples outside the dynamic range are excluded. %The standard deviation is chosen to reliably sample within the truncated range; that is $\sigma_\theta = 0.08$, $\sigma_{\dot{\theta}} = 0.2$ where $\sigma$ is the standard deviation. 
\item \textbf{Trajectories}: These datasets are composed of trajectories, beginning from 100 initial conditions that are simulated forward the same number of time steps. The dynamic range of this dataset differs from the two other dataset types.
\end{enumerate}

The models constructed for DDE and EDMD use the same observable functions. The observable functions chosen are two dimensional radial basis functions (RBFs), uniformly distributed between the maximum and minimum values of each state variable in their respective dataset, and the state variables. The total order of the system is 27th order with 25 RBFs and 2 state variables. 

\begin{figure} 
    \centering
    \includegraphics[width= 0.9\linewidth]{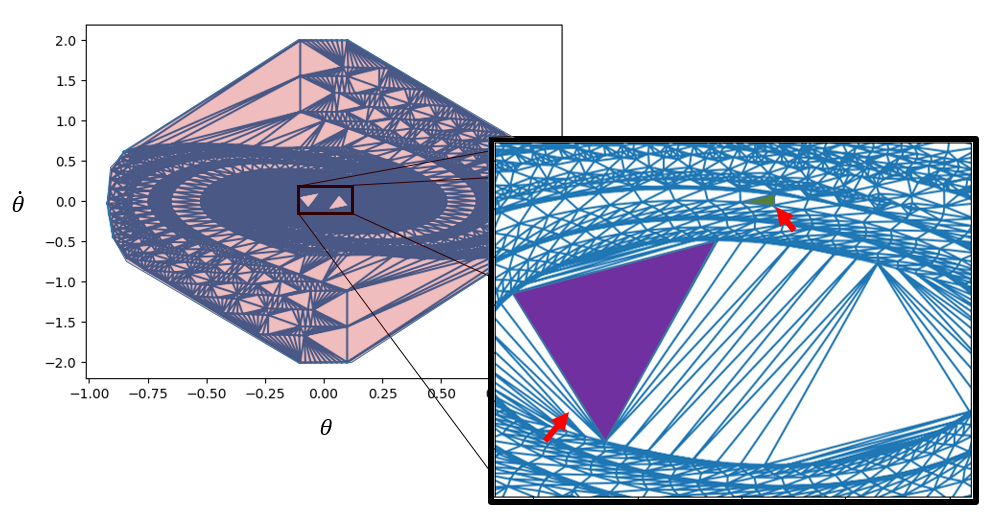}
    \caption{Graph of connections for a trajectory dataset composed of 10000 points formed when utilizing the data-driven direct encoding method. The lightly red shaded region denotes the dynamic range. In the zoomed-in image, the difference in volumes associated to different data points can be observed. Noting the difference in size of the purple triangle versus the green triangle.}
    \label{fig:pend_graph}
\end{figure}

A trajectory dataset graph is generated using Delaunay Triangles in DDE, shown in Fig. \ref{fig:pend_graph}. 
\begin{table}  
\caption{Sum of Squared Errors over dynamic range with varying dataset sizes.}
\label{tab:pred_dynrange} \begin{center} \begin{tabular}{c c c}
\hline \hline 

 \multirow{ 2}{*}{Dataset Size} & Total SSE & SSE Variance\\
 & EDMD / DDE  & EDMD / DDE\\ 
\hline 
\hline
\multicolumn{3}{c }{\textit{Uniform Datasets}} \\
\hline
\hline
900  & 19.470 / \textbf{17.167}  &  \textbf{0.0094} / 0.0097\\
2500 & 17.995 / \textbf{16.471} &  \textbf{0.0095} / 0.0103\\
10000 & 17.010 / \textbf{16.184} & \textbf{0.0099} / 0.0105\\
22500 & 16.698 / \textbf{16.133}  &  \textbf{0.0101} / 0.0105\\
\hline
\hline
% \multicolumn{4}{c}{\textit{Gaussian Datasets}} \\
% \hline
% \hline
% 1000  & 25.565$\pm$0.258 / \textbf{24.377$\pm$0.362} &  0.0135 / \textbf{0.0120}\\
% 2500 & 24.991$\pm$0.286 / \textbf{23.739$\pm$0.222}&  0.0139 / \textbf{0.0115}\\
%  5000 & 24.662$\pm$0.185 / \textbf{23.518$\pm$0.418} &  0.0142 / \textbf{0.0123}\\
% 10000 & 24.395$\pm$0.331 / \textbf{23.085$\pm$0.377} &  0.0155 / \textbf{0.0113}\\
% 25000 & 25.219$\pm$0.412 / \textbf{22.167$\pm$0.424} &  0.0156 / \textbf{0.0114}\\
% \hline
% \hline
\multicolumn{3}{c}{\textit{Trajectory Datasets}} \\
\hline
\hline
 1000  & 56.532 / \textbf{33.392} & 0.0349 / \textbf{0.0193}\\
% \hline
% EDMD & 1500  & 30.225  & 0.158\\
% DDE & 1500 & \textbf{24.788}  & \textbf{0.160}\\
2500 & 33.330 / \textbf{25.064} &  0.0200 / \textbf{0.0130}\\
5000 & 31.690 / \textbf{25.099} &  0.0195 / \textbf{0.0129}\\
10000 & 30.184 / \textbf{25.101} &  0.0186 / \textbf{0.0129}\\
25000 & 29.380 / \textbf{25.106} &  0.0181 / \textbf{0.0129}\\
\hline
\hline
\end{tabular} \end{center} \end{table}

\begin{figure} 
    \centering
    \includegraphics[width = 1\linewidth]{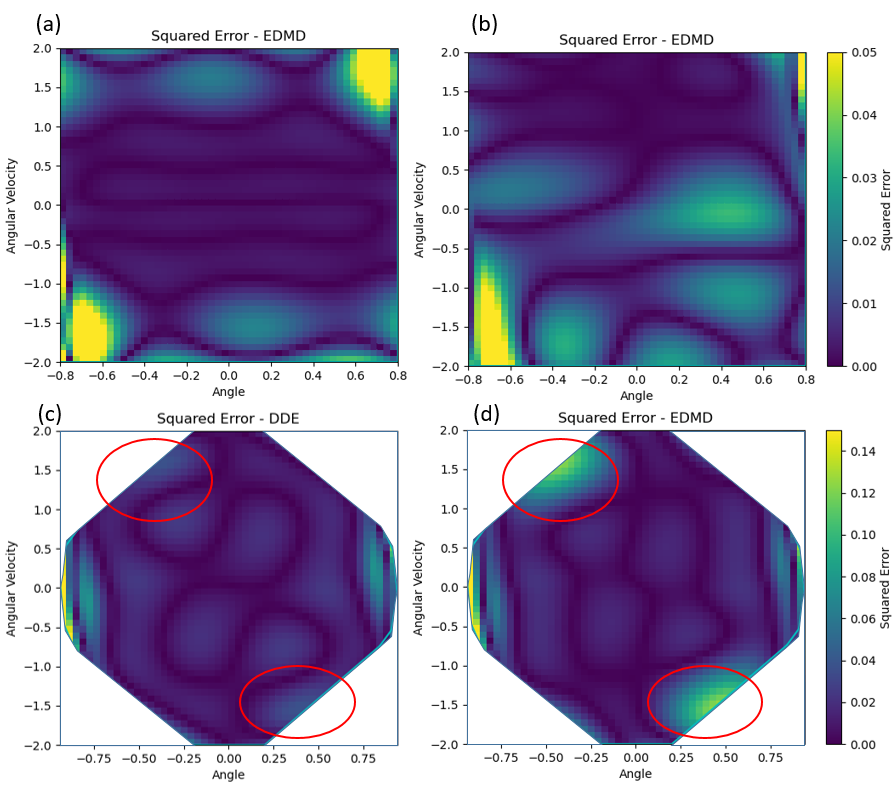}
    \caption{Sum of Squared Error plots for various datasets. (a) and (b) are EDMD models for Gaussian datasets; (a) uses a dataset that is centered at $[0,0]$ and (b) uses a dataset that is centered at $[0,2]$. (c) DDE and (d) EDMD models using the trajectory dataset composed of 2500 data points and using 25 RBFs. Circled in red are the regions with greatest variation in SSE between models. Only the dynamic range is shown in all plots.}
    \label{fig:sse_space}
\end{figure}

The accuracy of the models is tested through calculating sum of squared errors (SSE) for one-step ahead predictions over the dynamic range of the datasets. These error values are calculated for a uniform grid of points, similar to that used in the uniform datasets. A visualization of the SSE is plotted in Fig. \ref{fig:sse_space}. The results of these calculations are shown in Table \ref{tab:pred_dynrange} and \ref{tab:pred_gaussian}. In the computation, the dynamic range is discretized, and the SSE value of each point is summed. For the Gaussian datasets, the test is run for eight iterations of each dataset to account for randomness and the average result is noted.
\begin{table}  
\caption{Sum of Squared Errors over dynamic range for Gaussian distributions with different centers. Centers of the distribution are noted above each column.}
\label{tab:pred_gaussian} \begin{center} \begin{tabular}{c c c c}
\hline \hline
 \multirow{ 3}{*}{Dataset Size} & \multicolumn{3}{c }{Total SSE} \\
 & Center: $[0,0]$ & Center: $[0.8, 0]$ & Center: $[0,2]$ \\
 & EDMD / DDE  & EDMD / DDE & EDMD / DDE\\ 
\hline \hline 
\multicolumn{4}{c}{\textit{Gaussian Datasets}} \\
\hline
\hline
1000  & 25.565 / \textbf{24.377} &  25.161 / \textbf{22.98} & 24.338 / \textbf{22.445}\\
2500 & 24.991 / \textbf{23.739}&  25.421 / \textbf{22.747} & 24.221 / \textbf{21.590}\\
 5000 & 24.662 / \textbf{23.518} &  26.805 / \textbf{22.415} & 23.914 / \textbf{21.195}\\
10000 & 24.395 / \textbf{23.085} &  28.424 / \textbf{21.825} & 25.770 / \textbf{21.052}\\
25000 & 25.219 / \textbf{22.167} &  30.788 / \textbf{21.687} & 26.941 / \textbf{20.704}\\
\hline \hline
\end{tabular} \end{center} \end{table}

\begin{table} [!ht]
\caption{Sum of Squared Errors over dynamic range with varying order of lifted linear models.}
\label{tab:pred_obs} \begin{center} \begin{tabular}{c c c}
\hline \hline 
 \multirow{ 2}{*}{\# Observables} & \multicolumn{2}{c}{Total SSE}\\
  & EDMD & DDE \\
% Model Method & \# Observables & Total SSE & Max SSE\\
\hline 
\hline
\multicolumn{3}{c}{\textit{Trajectory Dataset, 5000 points}} \\
\hline
27  & 31.690 & \textbf{25.099} \\
51  & 36.657 & \textbf{21.637}\\
83 & 28.437 & \textbf{13.613}  \\
\hline
\end{tabular} \end{center} \end{table}

% \begin{figure}
%     \centering
%     \includegraphics[width= \linewidth]{Figures/pendulum/sse_fig_smalldata.png}
%     \caption{Sum of Squared Error (SSE) plot for (a) EDMD and (b) DDE models using the trajectory dataset composed of 2500 data points.}
%     \label{fig:trajpend_sse}
% \end{figure}

% \begin{figure}
%     \centering
%     \includegraphics[width= 0.8 \linewidth]{Figures/pendulum/gridpoint/graph.png}
%     \caption{Graph of connections for the grid point dataset formed when utilizing the data-driven direct encoding method.}
%     \label{fig:pend_gridgraph}
% \end{figure}

\subsection{Observable Function Variation}

The second experiment varies the number of observable functions selected, thus increasing the order. In this experiment, the number of RBFs is varied through uniformly increasing the density of the centers of the function over the range of the dataset. The results are noted in Table \ref{tab:pred_obs}. In the table, the number detailing number of observables is the number of functions including the state variables.

\subsection{Discussion}
From these results we can make the following observations.

\begin{figure}
    \centering
    \includegraphics[width = 0.85\linewidth]{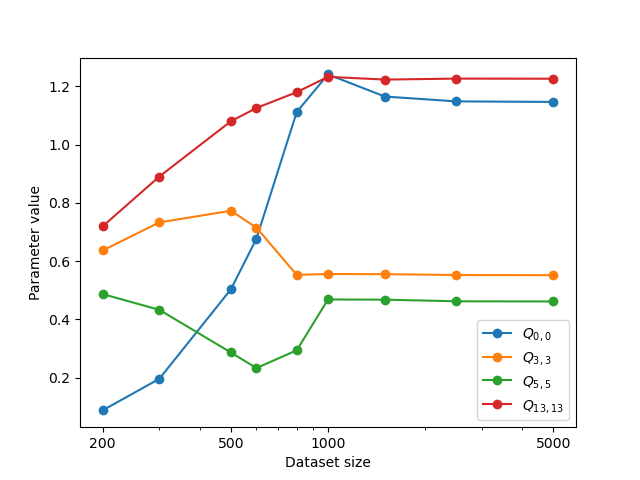}
    \caption{Change in values of diagonal elements in the $Q$ matrix of DDE with respect to trajectory dataset size. The specific elements shown correspond to a state variable, $Q_{0,0}$, and three RBF functions of varying distances away from the equilibrium.}
    \label{fig:qplot}
\end{figure}

\begin{itemize}
    \item All the numerical experiments show that DDE outperforms EDMD in total SSE.
    \item For uniform datasets, both models are nearly equivalent, though DDE has slightly lower SSE in all cases, shown in Table \ref{tab:pred_dynrange}. This result is expected as all data points are weighted equally in a uniform distribution.
    \item EDMD models exhibit significantly different distributions of prediction error, depending on dataset distribution. In Fig. \ref{fig:sse_space}-(a), the EDMD model learned from a dataset with high density near the origin produces a prediction error distribution that is low in accuracy in the top-right and the bottom-left corners of the dynamic range. In contrast, Fig.\ref{fig:sse_space}-(b)  illustrates that when using EDMD to learn from data with high density at $ \theta = 0, \dot{\theta} = 2$, the model results in low accuracy in the lower half of the dynamic range. DDE does not exhibit this high distribution dependency and achieves lower total SSE, as shown in Table \ref{tab:pred_gaussian}.
    \item In the trajectory datasets in Table \ref{tab:pred_dynrange}, DDE is not only lower in total SSE than EDMD but is also smaller in variance. Fig. \ref{fig:sse_space}-(c) shows that DDE has a uniformly low error distribution across the dynamic range, while EDMD in Fig.\ref{fig:sse_space}-(d) has two regions, as circled in the figure, with significantly larger error. %In the figure, the error plots over the dynamic range of the data are similar except for, as circled in the figure, two regions on which have a significant difference between models. 
    These regions coincide with sparsity in the dataset, providing evidence of EDMD's bias towards regions of high data density and explaining the difference in performance between models for the non-uniform datasets.
    \item In the trajectory datasets, the total SSE converges for DDE beginning with small dataset sizes. This result implies that the elements in the $R$ and $Q$ matrices of DDE, that is, the inner product integral computations, converge as the data size and the data density increase. This convergence is confirmed in Fig. \ref{fig:qplot}, where several elements of the $Q$ matrix are plotted against the data size. 
    \item The second experiment, regarding variations in observable function numbers demonstrates the effect of DDE remains even for significant increases in observable functions, referring to Table \ref{tab:pred_obs}. In all cases tested, DDE significantly outperforms EDMD over the dynamic range, as expected for a non-uniform dataset.
\end{itemize}

\section{Application to Neural Net Koopman Modeling}
\label{nnkoop}
The use of deep neural networks for finding effective observable functions and constructing a Koopman linear model has been reported by several groups \cite{Lusch_2018, yeung2017, Selby_2021}. This method, sometimes referred to as Deep Koopman, is effective for approximating the Koopman operator to a low-order model, compared to the use of locally activated functions, such as RBFs, which scale poorly for high-order nonlinear systems. The proposed DDE method can be incorporated into Deep Koopman, further improving approximation accuracy.

\begin{figure}
    \centering
    \includegraphics[width = 0.8\linewidth]{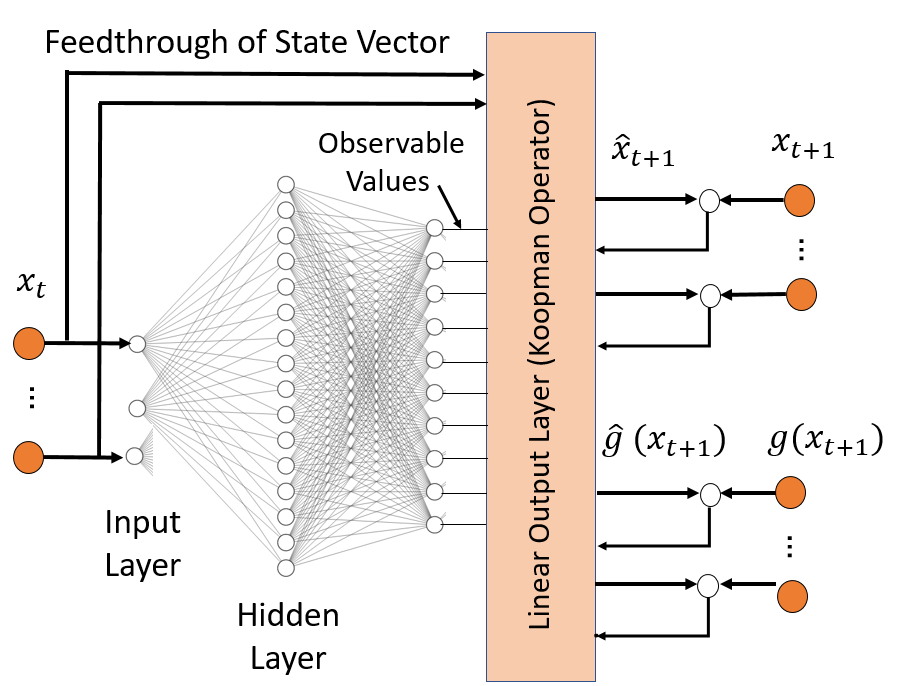}
    \caption{Feedforward neural network model used to generate observable functions. The final layer of the neural network is a linear layer. In the standard Deep Koopman model this remains the same after the model is fully trained. However, with the DDE model, this final layer is recalculated using DDE by taking in the dataset used to train the model.}
    \label{fig:nn}
\end{figure}

Fig. \ref{fig:nn} shows the architecture of the neural network similar to the prior works \cite{Lusch_2018, yeung2017, Selby_2021}. The input layer receives training data of independent state variables. The successive hidden layers produce observable functions; these functions feed into the output layer consisting of linear activation functions. This linear output layer corresponds to the $A$ matrix that maps the observables of the current time to those of the next time step, i.e. the state transition in the lifted space. In the Deep Koopman approach, the output layer, that is, the $A$ matrix, is trained together with observable functions in the hidden layers. Using the observable functions learned from deep learning, the output layer weights are replaced with the $A$ matrix obtained from DDE, captioned in Fig. \ref{fig:nn}.

%(EBP). EBP is basically a local-gradient based minimization of a loss function, which inevitably faces numerous challenges, including local minima, over-fitting, and hyper-parameter tuning. There are many techniques to get rid of these difficulties, but it requires a significant amount of trial-an-error effort. The deep neural network training alone is limited as a methodology for constructing a Koopman model.

 %The number of the units in the final hidden layer connected to the linear output layer is the number of observables. %For this set of observable functions and data matrices $D_N$ and $D_N^f$, we can apply the DDE algorithm for finding the A matrix.
%The neural net training can be repeated for refining the observables after the A matrix is determined from the DDE. Furthermore, the DDE and the neural net training can be repeated multiple times in a bootstrap manner.

% The trained observable functions are used for constructing a high-dimensional state vector $z$. %: 
% \begin{equation}
%     z = [g_1(x; w_1), \cdots, g_m(x; w_m)]^T  
% \end{equation}
% where $w_k$ is the weights of all the hidden layer units associated to the $k$-th state variable in the lifted space. 
The effectiveness of this Deep Koopman-DDE method is applied to a multi-cable manipulation system \cite{ng2021model}. A simplified single cable version is utilized consisting of six independent state variables. The network is constructed using PyTorch, trained with an Adam optimizer, to generate 40 observable functions. These model concatenates the state variables of the nonlinear system, resulting in a 46th order model. Table \ref{tab:nndef} shows parameters used for training the Deep Koopman model. The training dataset is composed of 3000 data points drawn from trajectories. Table \ref{table:SSE of Neural Net} compares the Deep Koopman model to the proposed model that uses DDE. Results are in terms of sum of squared error over a set of test trajectories. A significant improvement is achieved by incorporating DDE into the deep learning method.
\begin{table} [!h]
    \caption{Neural network parameters and characteristics.}
    \small
    \centering
    \begin{tabular}{c|c}
    \hline
        Parameters and Characteristics & Value \\
        \hline
        Number of Hidden layers & 3 \\
        Activation Functions, Both Hidden Layers & ReLU \\
         Width of 1st Hidden Layer & 16 \\
         Width of 2nd Hidden Layer & 16 \\
         Width of 3rd Hidden Layer & 40 \\
         Learning Rate, $\alpha$ & 0.01 \\
    \end{tabular}
    \label{tab:nndef}
\end{table}
\begin{table} [!h]
\caption{Average SSE prediction error for trajectories in set of test data for single winch system.}
\label{table:SSE of Neural Net} \begin{center} \begin{tabular}{c c c}
\hline \hline 
Modeling Method & 1 Time Step & 20 Time Steps  \\
\hline
\hline
Deep Koopman only  & 0.2471 & 9.8610  \\
Deep Koopman + DDE  & \textbf{0.2350} & \textbf{4.1131}  \\
\hline
\end{tabular} \end{center} \end{table}

Practical concerns arose from the use of Delaunay Triangulation in this experiment. Specifically, the method was found to be limiting and inconsistent for systems that are eighth order or higher, due to computational cost and numerical instability. Alternative numerical integration approaches or alternative partitioning methods may be necessary when applying the method to higher order nonlinear systems. 
\section{Conclusion}
\label{conclusion}

In this work,  a new data-driven approach to generating a Koopman  linear model based on the direct encoding of Koopman Operator (DDE) is presented as an alternative to dynamic mode decomposition (DMD) and other related methods using least squares estimate (LSE). The major contributions include: 1) The analytical formula of Direct Encoding is converted to a numerical formula for computing the inner product integrals from given data; 2) An efficient algorithm is developed and its convergence conditions to the true results are analyzed; 3) Numerical experiments demonstrate a) greater accuracy compared to EDMD, b) lower sensitivity to data distribution, and c) rapid convergence of inner product computation. Furthermore, the DDE method is incorporated to Deep Koopman, i.e. neural network based methods for construction of the Koopman Operator, for improving prediction accuracy. %The authors justify the use of DE based methods over DMD based methods and utilize numerical experiments that demonstrate:
%1) greater accuracy over the dynamic range defined by the dataset used for training, 2) convergence of the DDE model with small sizes of datasets relative to system order, 3) improvements 

%While the method presented is useful for prediction accuracy, we offer future directions that can be pursued to improve the applicability of the DDE method to real world scenarios. One dimension of improvement is the creation of an update method. For least squares based models, recursive least squares serves as a reasonable method for incorporation of new data points to update the model. However, because the model is an integral-based model that utilizes a graph for numerical integration, an algorithm must be developed to quickly incorporate the new data point into the graph and update the linear model. Another area worth pursuing is in incorporation of control to the method. 
%However, in the current formulation, DE and DDE are only viable for autonomous systems, as incorporation of control raises concerns. Should a formulation be created that assuages these issues, then the usefulness of the method increases drastically, analogous to EDMD with control. 
% \appendix
% \input{appendix}
\bibliographystyle{IEEEtran}
\bibliography{IEEEabrv, main.bib}
\end{document}